%% file: paper.tex
\documentclass[letterpaper, 10 pt, conference]{ieeeconf}  

\IEEEoverridecommandlockouts                              
 \pdfoutput=1 
\input{packages}
\input{commands}

\author{Antony Thomas$^\ast$ and Sunny Amatya$^\dagger$ and Fulvio Mastrogiovanni$^\ast$ and Marco Baglietto$^\ast$
\thanks{$^\ast$Department of Informatics, Bioengineering, Robotics, and Systems Engineering, University of Genoa, Via All'Opera Pia 13, 16145 Genoa, Italy. \textit{
antony.thomas@dibris.unige.it, fulvio.mastrogiovanni@unige.it,  marco.baglietto@unige.it}}%
\thanks{$^\dagger$Bio-Inspired Mechatronics Lab, Arizona State University at the Polytechnic Campus, Mesa, Arizona 85212. \textit{samatya@asu.edu}}%
        }

\title{Task-assisted Motion Planning in Partially Observable Domains}

\begin{document}
\maketitle
\begin{abstract}
We present an integrated Task-Motion Planning framework for robot navigation in belief space. Autonomous robots operating in real world complex scenarios require planning in the discrete (task) space and the continuous (motion) space. To this end, we propose a framework for integrating belief space reasoning within a hybrid task planner. The expressive power of PDDL+ combined with heuristic-driven semantic attachments performs the propagated and posterior belief estimates while planning. The underlying methodology for the development of the combined hybrid planner is discussed, providing suggestions for improvements and future work. Furthermore we validate key aspects of our approach using a realistic scenario in simulation.
\end{abstract}

\section{Introduction}
\input{intro}
\label{intro}

\section{Related Work}
\input{related_work}

\section{Task-Motion Planning: Preliminaries and Notations}
\input{problem_definition}

\section{TMP Design and Implementation}
\input{methodology}

\section{Empirical Evaluation}
\input{results}

\section{Conclusion}
\input{conclusion}

\bibliographystyle{plain}
\bibliography{/home/antony/Research_Genoa/References/References}

\end{document}

%% file: packages.tex
\usepackage[english]{babel}
\usepackage[utf8]{inputenc}

\usepackage{epsfig} 
\usepackage{mathptmx} 
\usepackage{times} 
\usepackage{amsmath} 
\usepackage{amssymb}  

\usepackage{color,xcolor,ucs}
\usepackage{xr-hyper}
\usepackage{subfig}
\usepackage{caption}

\usepackage{floatrow}
\usepackage{tabularx}
\usepackage{float}
\usepackage{amsfonts}

\usepackage{amsthm}
\usepackage{makeidx}         
\usepackage{comment}         
\usepackage{graphicx}        
\usepackage{multicol}        
\usepackage[bottom]{footmisc}
\usepackage{bm}

\usepackage{url}

\usepackage{xr-hyper}

\usepackage{algorithm}
\usepackage{algpseudocode}

\usepackage{linegoal}

\usepackage{xspace}
\usepackage{rotating}

\usepackage{tikz}
\usepackage{tkz-graph}
\usetikzlibrary{arrows,shapes,shadows,positioning,calc}

\usepackage{graphicx}
\usepackage{threeparttable}
\usepackage{multirow}
\usepackage[font=scriptsize,labelfont=bf]{caption}

\usepackage{listings}

\usepackage{wrapfig}

%% file: commands.tex
\makeatletter
\algnewcommand{\LineComment}[1]{\Statex \hskip\ALG@thistlm \(\triangleright\) #1}
\algnewcommand{\LineCommentCont}[1]{\Statex \hskip\ALG@thistlm \parbox[t]{\linegoal}{\hangindent=1em\hangafter=1 $\triangleright$ #1}}
\makeatother

\newsavebox{\mybox} 

\theoremstyle{remark}
\newtheorem{defn}{Definition}

%% file: intro.tex
Autonomous robots operating in complex real world scenarios require different levels of planning to execute their tasks. High-level (task) planning helps break down a given set of tasks into a sequence of sub-tasks, actual execution of each of these sub-tasks would require low-level control actions to generate appropriate robot motions. In fact, the dependency between logical and geometrical aspects is pervasive in both task planning and execution. Hence, planning should be performed in the task-motion or the discrete-continuous space. 

In recent years, combining high-level task planning with low-level motion planning has been a subject of great interest among the Robotics and Artificial Intelligence (AI) community. Traditionally, task planning and motion planning have evolved as two independent fields. AI planning frameworks as the Planning Domain Definition Language (PDDL)~\cite{mcdermott1998AIPS} mainly focus on high-level task planning supposing that the geometric preconditions (e.g., grasping poses for a pick-up task~\cite{srivastava2014ICRA}) for the robot motion to carry out these tasks are achievable. However, in reality, such an assumption can be catastrophic as an action or sequence of actions generated by the task planner might turn out to be unfeasible at the controller execution level.

Let us consider a toy scenario in which a robot needs to move from location A to location B. At the task level this corresponds to an action ``\textit{goto A B}'', taking the robot from A to B. The actual execution of the task requires motion planning to find a suitable collision free path between A and B. At the task level there seems to be no harm in executing the \textit{goto} action. Yet, at the motion planning level it might happen that A and B have no connected paths joining them. This renders the task action infeasible. Now also suppose that there is a constraint on the robot battery consumption. In such a scenario, a motion plan alone will not suffice as it is required to logically reason to determine the actions that lead to minimal battery consumption. Though a simple scenario, it clearly illustrates the need for a combined Task-Motion Planning (TMP) strategy.



Real-world scenarios often induce uncertainties. Such uncertainties arise due to insufficient knowledge about the environment, inexact robot motion or imperfect sensing. In such scenarios, the robot poses or other variables of interest can only be dealt with, in terms of probabilities. Planning is therefore done in the \textit{belief} space, which corresponds to the probability distributions over possible robot states. Consequently, for efficient planning and decision making, it is required to reason about future belief distributions due to candidate actions and the corresponding expected observations. Such a problem falls under the category of Partially Observable Markov Decision Processes (POMDPs)~\cite{kaelbling1998AI}. Hence, in a TMP approach, the task planner should be capable of reasoning in the belief space while synthesizing a plan. Besides, the task planner also requires some amount of geometrical information about the
environment and the robot itself, the lack of which may lead to
undesirable plans. At the execution level, the motion planner might encounter unexpected scenarios notwithstanding
the plan provided. This calls for a re-plan, updating the task
planner with the new belief, resulting in a cyclic interdependency. Consequently, both task and motion planning are interdependent and should not be considered as separate processes.

This paper presents an extension of the work in~\cite{thomas2018LTA} and provides a more comprehensive evaluation of the approach presented. The main contributions are as follows: (1) developing an integrated TMP algorithm for mobile robot planning in the belief space. It is to be noted that we are concerned with the problem of TMP for navigation unlike the popularly investigated problem of TMP for manipulation. (2) Exploiting the expressive power of PDDL+~\cite{fox2006JAIR} to simulate robot motions and perform the belief estimates within PDDL+. To the best of our knowledge, this is the first TMP approach based on PDDL+ planning semantics. (3) Our domain description can hence be employed for any mobile robot planning problem in general. 


%% file: related_work.tex

The genesis of TMP can be credited to Fikes and Nilsson for their work on STRIPS~\cite{fikes1971strips} which further led to the Shakey project~\cite{nilsson1984shakey}. Shakey's planner performed a logical search first, assuming that the resulting robot motion plans can be formulated. This assumption limits the capability of the agent as the high-level actions may turn out to be non executable due to geometric limitations. Later works either carried out the generated plans, validating them using a robot motion planner~\cite{dornhege2009SSRR} or performed a combined search in the logical and geometric spaces using a state composed of both the symbolic and geometric paths~\cite{cambon2009IJRR}. The aSyMov planner used in~\cite{cambon2009IJRR} adopts a combination of Metric-FF~\cite{hoffmann2003JAIR} and a sampling based motion planner. In contrast, we use a hybrid temporal task planner~\cite{piotrowski2016AAAI} incorporating robot state uncertainty. Srivastava \textit{et al.}~\cite{srivastava2014ICRA} implicitly incorporate geometric variables, performing symbolic-geometric mapping using a planner-independent interface layer.



Kaelbling and Lozano-P\'{e}res~\cite{kaelbling2012aTR} propose a hierarchical approach that tightly integrates the logical and geometric planning. The complexities arising out of long horizon planning are tackled to the extent that planning is done at different levels of abstraction, thereby reducing the long horizons to a number of feasible sub-plans of shorter horizon. This regression-based planner assumes that the actions are reversible while backtracking. In contrast to their earlier work the serializability assumption of the subgoals is relaxed. This work is extended in~\cite{kaelbling2013IJRR} to consider the current state uncertainty, modeling the planning problem in the belief space. Uncertain outcomes are modeled by converting  a Markov decision processes (MDP) into a weighted graph, thereby modifying their earlier approach of \textit{hierarchical planning in the now}. Belief update is then performed when observations are obtained. Phiquepal and Toussaint~\cite{phiquepal2017RSSws} discuss an ongoing work for TMP under partial observability, computing long-horizon policies that are arborescent in nature.

The above discussed approaches focus on finding feasible plans sacrificing optimality, emphasizing on performance. Toussaint~\cite{toussaint2015IJCAI} performs optimization over an objective function based on the final geometric configuration (and the cost thereby), finding approximately locally optimal solutions by minimizing the objective function. The planning problem is modeled as a constraint satisfaction problem with symbolic states used to define the constraints in the optimization. Lozano-P\'{e}res and Kaelbling~\cite{lozano2014IROS} model the motion planning as a constraint satisfaction problem over a subset of the configuration space. Iteratively Deepened Task and Motion Planning (IDTMP) is a constraint based task planning approach that incorporates geometric information (motion feasibility) at the task planning level~\cite{dantam2018IJRR}. In our architecture, the waypoints fed into the task planner are generated using the motion planner, similar to the motion planner information that guides the IDTMP task planner. IDTMP performs task-motion interaction using abstraction and refinement functions whereas we use \textit{Semantic attachments}~\cite{dornhege2009ICAPS} to that aim.

FFRob~\cite{garrett2018IJRR} performs task planning by performing search over a sampled finite set of poses, grasps and configurations. In this way, a motion planner is not required as the task planning is performed directly over the sampled set. We also directly perform task planning over a set of sampled poses, implicitly generating motion plans for each task-level actions. PETLON~\cite{lo2018AAMAS} is the work closest to our approach since they also discuss a TMP approach for navigation that is task-level optimal. However, the action costs returned by their motion planner is the trajectory length and they assume completely observable domains. In contrast, we consider a partially observable environment and the costs associated with the uncertainties arising out of the same.

%% file: problem_definition.tex
TMP essentially involves combining discrete and continuous decision-making to facilitate efficient interaction between the two domains. Below we define the TMP problem formally. 

\begin{defn}\textit{Task} domain can be represented via state transition system and is a tuple $\Sigma = (S, A, \gamma, s_0, S_g)$ where
\end{defn}
\begin{itemize}
\item $S$ is a finite set of states
\item $A$ is a finite set of actions
\item $\gamma : S \times A \rightarrow S$ is the state transition function such that $s' = \gamma(s, a)$
\item $s_0 \in S$ is the start state
\item $S_g \subseteq S$ is the set of goal states
\end{itemize}

\begin{defn}\textit{Task} Plan for a task domain $\Sigma$ is the sequence of actions $a_0,...,a_n$ such that $s_{i+1} = \gamma(s_i, a_i)$, for $i = 0,...,n$ and $s_{n+1}$ satisfies $S_g$.
\end{defn}
 
Due to the popularity of the Planning Domain Definition Language (PDDL)~\cite{mcdermott1998AIPS} among the Planning community, we resort to the same for modeling our task domain. 

\begin{defn}\textit{Motion} Planning Problem is a tuple $M = (C, f, q_0, G)$ where
\end{defn}
\begin{itemize}
\item $C$ is the configuration space
\item $f =\{1,0\}$ determines if a configuration is collision free ($C_{free}$ with $f =1$) or not ($f=0$)  
\item $q_0$ is the initial configuration
\item $G$ is the set of goal configurations
\end{itemize}

A motion plan essentially involves finding a valid trajectory in $C$ from $q_0$ to $q_n \in G$ such that $f$ evaluates to true for $q_0,...,q_n$. A motion plan can also be defined as $\tau : [0, 1] \rightarrow C_{free}$ such that $\tau(0) = q_0$ and $\tau(1) \in G$. We will use a combination of the two to define the TMP problem and use a Rapidly-exploring Random Tree (RRT)~\cite{kuffner2000ICRA} based sampling strategy to generate collision free configurations.  

\begin{defn}\textit{Task-Motion} Planning Problem is a tuple $(C, M, \phi, \xi)$ where
\end{defn}
\begin{itemize}
\item $\phi : S  \rightarrow 2^ C$, mapping states to the configuration space 
\item $\xi : A  \rightarrow 2^ C$, mapping actions to motion plans
\end{itemize}
and the TMP problem is to find a sequence of actions $a_0,...,a_n$ such that $s_{i+1} = \gamma(s_i, a_i)$, $s_{n+1} \in S_g$ and to find a sequence of motion plans $\tau_0,...,\tau_n$, $\tau_n(1) \in G$ such that for $i = 0,...,n$, it holds that

\vspace{-0.5cm}

\begin{align}
& \tau_i(0) \in \phi(s_i) \ \textrm{and} \ \tau_i(1)  \in \phi(s_{i+1})  \\
&\tau_{i+1}(0) = \tau_i(1)   \\
&\tau_i \in \xi(a_i)
\end{align}

In this paper, we consider the TMP for navigation of a mobile robot operating in a pre-mapped environment.
At any time $k$, we denote the robot pose (or configuration $q_k$) by $x_k\doteq(x, y, \theta)$, the measurement acquired is denoted by $z_k$ and the control action applied is denoted as $u_k$. The robot kinematics is modeled using the standard odometry based motion model

\vspace{-0.45cm}
\begin{equation}
\begin{split}
x' & = x + \delta_{trans} \cdot \cos(\theta+ \delta_{rot1})\\
y' & = y + \delta_{trans} \cdot \sin(\theta+ \delta_{rot1})\\
\theta' & = \theta + \delta_{rot1}+ \delta_{rot2}
\end{split}
\label{eq:odometry_model}
\end{equation}

\noindent
where $x_{k+1}\doteq(x', y', \theta')$ and $u_k \doteq (\delta_{rot1}, \delta_{trans}, \delta_{rot2})$ is the control applied (motion plan $\tau_k$). For brevity we write (\ref{eq:odometry_model}) as $x_{k+1} = f(x_k,u_k)+ \ w_k \sim \mathcal{N}(0,R_k)$, where $w_k$ is the zero-mean Gaussian noise.

%
To process the landmarks in the environment we measure the range and the bearing of each landmark relative to the robot's local coordinate frame, which can be specifically written as

\vspace{-0.45cm}
\begin{equation}
z_k  = \begin{bmatrix} 
r    \\
\phi 
\end{bmatrix} + v_k \
,  \ v_k \sim \mathcal{N}(0,Q_k)
\label{eq:measurement_model}
\end{equation}
 

\noindent
where $r$, $\phi$ are the range and bearing respectively and $v_k$ the zero-mean Gaussian noise. For brevity, (\ref{eq:measurement_model}) will be written as $z_{k} = h(x_k,lm_i) +\ v_k \sim \mathcal{N}(0,Q_k)$. It is to be noted that we assume data association as solved and hence given a measurement we know the corresponding landmark that generated it. It is possible to relax this assumption to incorporate reasoning regarding data association within the belief space, as shown recently in~\cite{pathak2018IJRR}.

The motion (\ref{eq:odometry_model}) and observation (\ref{eq:measurement_model}) models can be written probabilistically as
$p(x_{k+1}|x_k, u_k)$ and $p(z_k|x_k)$ respectively. Given an initial distribution $p(x_0)$, and the motion and observation models, the posterior probability distribution at time $k$ can be written as

\vspace{-0.6cm}
\begin{equation}
p(X_k|Z_k, U_{k-1}) = p(x_0)\prod_{i=1}^k p(x_{k}|x_{k-1}, u_{k-1}) p(z_k|x_k)
\end{equation}
where $X_{0:k} \doteq \{x_0,...,x_k\}$, $Z_{0:k}  \doteq\{z_0,...,z_k\}$ and $U_{0:k-1} \doteq \{u_0,...,u_{k-1}\}$. This posterior probability distribution is the \textit{belief} at time $k$, denoted by $b[X_k] \sim \mathcal{N} (\mu_k, \Sigma_k)$. Similarly, given an action $u_k$, the propagated belief can be written as

\vspace{-0.3cm}
\begin{equation}
b[\bar{X_{k+1}}] = p(X_k|Z_k, U_{k-1})p(x_{k+1}|x_k, u_k)
\end{equation}

Given the current belief $b[X_k]$, the control $u_k$, the propagated belief parameters can be computed using the standard Extended Kalman Filter (EKF) prediction as 

\vspace{-0.5cm}
\begin{equation}
\begin{split}
\bar{\mu}_{k+1} & = f(\mu_k, u_k)\\
\bar{\Sigma}_{k+1}   & = F_{k} \Sigma_k F_{k}^T + R_k
\end{split}
\label{eq:predict}
\end{equation}
where $F_k$ is the Jacobian of $f(\cdot)$ with respect to $x_k$. Upon receiving a measurement $z_k$, the posterior belief $b[X_{k+1}]$ is computed using the EKF update equations

\vspace{-0.4cm}
\begin{equation}
\begin{split}
K_k     & = \bar{\Sigma}_{k+1} H_k^T(H_k \bar{\Sigma}_{k+1}  H_k^T + Q_k)^{-1}\\
\mu_{k+1} & = \bar{\mu}_{k+1} + K_k(z_{k+1}-h(\bar{\mu}_{k+1},l_i))\\
\Sigma_{k+1} & = (I -K_k H_k)\bar{\Sigma}_{k+1} 
\end{split}
\label{eq:update}
\end{equation}

\noindent
where $H_k$ is the Jacobian of $h(\cdot)$ with respect to $x$, $K_k$ is the Kalman gain and $\mathbb{I} \in \mathbb{R}^{3 \times 3}$.

%% file: methodology.tex
In this Section we detail our TMP planner concept and approach. We begin by making the following observation. Planning in the belief space to obtain an optimal control policy essentially requires synthesizing a sequence of actions that minimize an application dependent objective function. Finding such an action sequence inherently involves searching in the motion space. Consequently, we employ task planning to perform this search. 
\begin{figure}[t!]
	\centering
	\caption{The TMP planner workflow.}
	\includegraphics[scale=0.2]{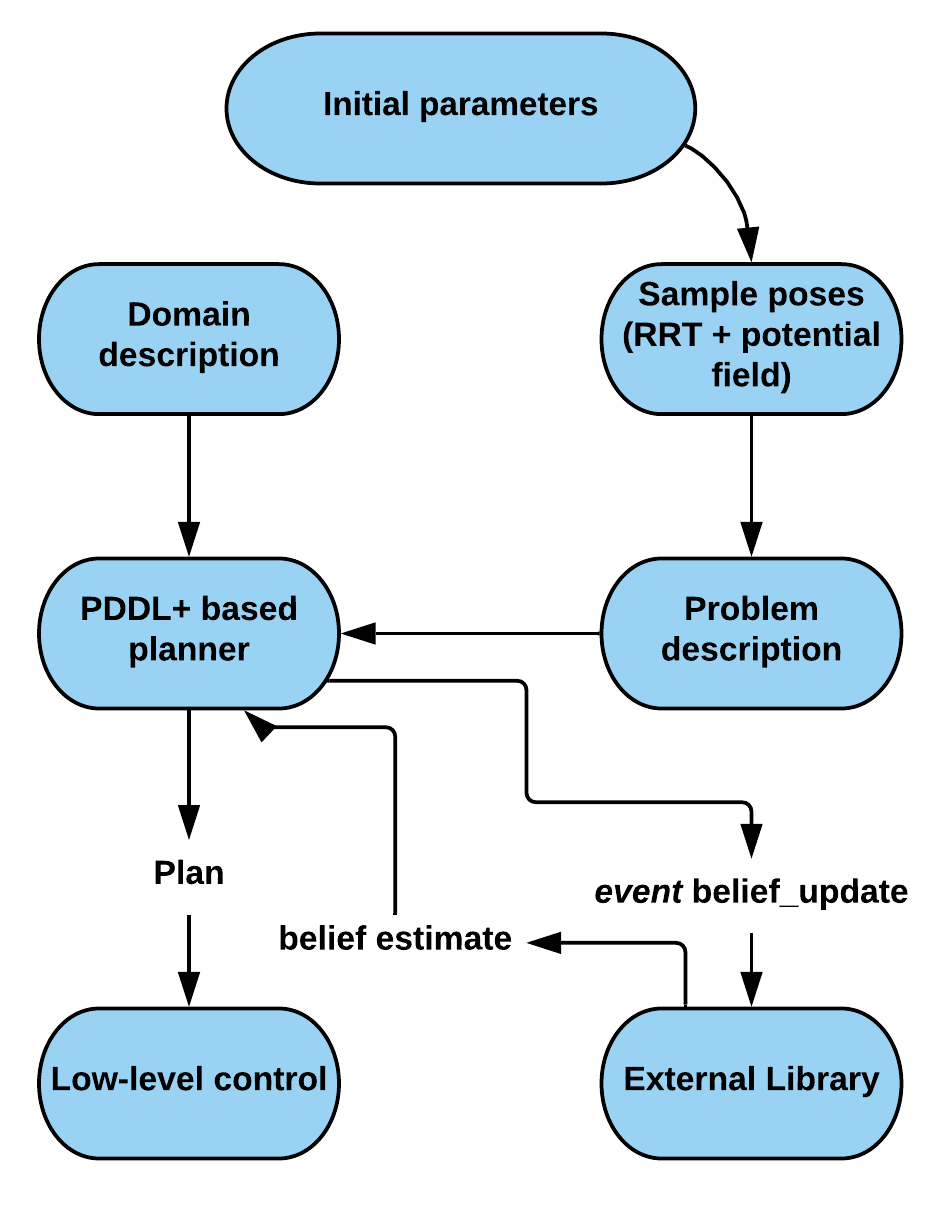}
	\label{fig:TMP_architecture}
\end{figure}

\subsection{Rationale and Scenario}

PDDL based planning frameworks are limited, as they are incapable of handling rigorous numerical calculations. Most approaches perform such calculations via external modules or \textit{semantic attachments}, e.g.~\cite{dornhege2009ICAPS}. The term semantic attachment was coined by Weyhrauch~\cite{weyhrauch1980AI} to describe attaching algorithms to function and predicate symbols via external procedures. Yet, the effects returned by these semantic attachments are not exploited in identifying \textit{helpful actions} during search and hence do not provide any heuristic guidance, deeming the task unsolvable most often. An action is considered \textit{helpful} if it achieves at least one of the lowest level goals in the relaxed plan to the state at hand~\cite{hoffmann2003JAIR}. Recently Bernardini \textit{et al.}~\cite{bernardini2017ICAPS} developed a PDDL based POPF-TIF planner to implicitly trigger such external calls via a specialized semantic attachments called \textit{external advisors}. They classify variables into direct, indirect and free. Direct (free) variables are the normal PDDL function variables whose values are changed in the action effects, in accordance with PDDL semantics. The indirect variables are affected by the changes in the direct variables. A change in a direct variable triggers the external advisor which in turn updates the indirect variables. POPF-TIF is based on the temporal extension of the metric-FF planner~\cite{hoffmann2003JAIR}. An intriguing feature of the planner is that it uses approximate values of the indirect variables at the Temporal Relaxed Plan Graph (TRPG) construction stage, incorporating these values during heuristic calculation thereby resulting in an efficient goal-directed search. During the forward state space search, the external advisor is called, updating the indirect variables with the exact values. 

Using semantic attachments that incorporate heuristic evaluation during the Relaxed Plan Graph (RPG) construction, we develop a hybrid planning framework capable of reasoning in the robot belief space while synthesizing a plan. We use PDDL+~\cite{fox2006JAIR} to model the planning task, providing the robot with a sequence of actions that can be passed on to the low-level controller for execution. PDDL+ provides the ability to model continuous temporal change via \textit{processes} and discrete exogenous activities in the environment via \textit{events}, thus relaxing the closed world assumption. The processes are similar to durative actions and the events are akin to instantaneous actions. However, processes and events are distinct from actions since a process or an event is triggered as soon as its precondition is satisfied whereas an action trigger depends on the planner search strategy. State uncertainty is incorporated in our model and synthesizing an efficient plan requires performing the belief updates within the task planner. PDDL+ \textit{processes} enable the simulation of robot motion with time and the \textit{events} are leveraged to perform the corresponding belief estimates. In our case, we use the DiNo planner~\cite{piotrowski2016AAAI} since it enables heuristic search for linear and non-linear systems using the entire set of PDDL+ features. 


\subsection{Task Description in PDDL+}

As discussed before, we consider a mobile robot in a known environment (i.e., map is given) with uncertainty in its initial pose. The set of landmarks in the environment are given by $lm = \{lm^1,...,lm^n\}$. The landmarks are features in the environment and are not to be confused with the landmarks in heuristic planning where they are intended as a set of operators such that each plan must contain some element of this set. The goal is to reach a certain final state $s_{n+1} \in S_g$ subject to minimizing an objective function

\vspace{-0.4cm}
\begin{equation}
J_k(u_{k:k+L-1}) \doteq \mathbb{E}\left\{ \sum_{l=0}^{L-1} c_l \left(
b[X_{k+l}], u_{k+l} \right) + c_L \left(
b[X_{k+L}] \right) \right\}
\label{eq:cost}
\end{equation}

\noindent
where $L$ is the look-ahead step, $c(\cdot)$ is the immediate cost for each look-step and the expectation is over the future observations, since these are unknown at the planning time. Our cost is a combination of distance to goal $c_G$ and the state uncertainty $c_{\Sigma}$ which is defined as $c_{\Sigma} = trace(\Sigma)$. DiNo performs a modified Enforced Hill-Climbing (mEHC) search starting from the initial state $s_0$, whereas, to absorb our cost function, we modify the search to a weighted $\textrm{A}^\star$ algorithm encompassing $c_{\Sigma}$ within $g(\cdot)$. Furthermore, to incorporate belief evolution while planning, the DiNo planner is extended to support external calls evaluating the belief at each planning stage. The belief, the process and the measurement noise are assumed to be Gaussian. 
 
DiNo tackles continuous nonlinear systems using the planning-via-discretization paradigm and hence a coarse planner discretization can lead to skipping certain decision points leading to invalid solutions. Most real world problems exhibit time constraints for decision making. The temporal planning horizon $T$ limits the plans to a maximum number of clock ticks, also making the states $S$ finite. Therefore, our TMP approach depends directly on the temporal planning horizon and the used PDDL+ process discretization. A longer horizon and shorter discretization increases planning complexity and directly affects the planning time. In Section~\ref{section:results}, we evaluate the performance based on these factors and analyze how our approach cope with the changes in these parameters.   

\subsection{Planner Workflow} 
\label{subsec:workflow}
An overview of our TMP planner framework is shown in Fig.~\ref{fig:TMP_architecture}. We assume that the environment map, the robot's initial belief $\mathcal{N}(\mu_0,\Sigma_0)$, the goal pose to be reached $x_g$ are known at the planning time. As discussed at the beginning of this Section, we utilize task planning to synthesize a plan, performing search in the motion space. We would like to reiterate the fact that in this paper we consider the problem of TMP for navigation and this requires synthesizing feasible configurations/paths for navigating to the goal(s). Once the map of the environment is known, collision-free configurations/poses are sampled. Task planning is performed over this sampled set of poses, generating a path to the goal. In this way, the task planner \textit{assists} in finding a feasible motion plan.

\begin{figure}[t!]
 \caption{Illustration for the RRT based potential field approach for sampling poses.}
\includegraphics[width=0.6\textwidth]{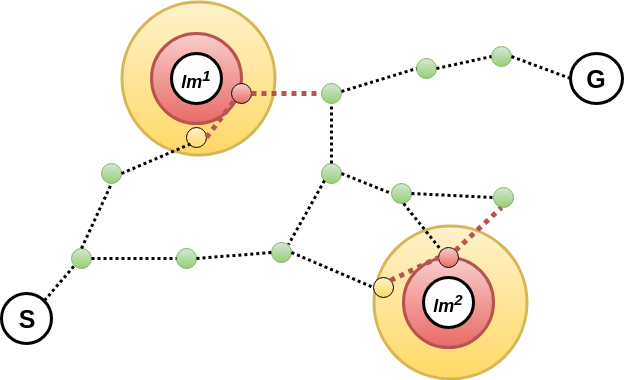}
\label{fig:SRRT}
\end{figure}

Standard RRT based approaches sample points connecting start and end locations. However, $c_{\Sigma}$ penalises the states with higher uncertainty and hence to minimize our objective function as given in (\ref{eq:cost}), it is to be ensured that such connected paths have ample number of points from which landmarks can be observed. Since the configuration space of the robot is the set of all poses, we sample poses while building the RRT. To facilitate this perception-aware search we implement an RRT based potential field approach to sample such relevant poses in the environment. This strategy is shown in Fig.~\ref{fig:SRRT}. poses near the potential field of the landmarks are pulled closer towards it (light red node in the Figure) and once a sufficient number (currently user defined) of such poses are generated, the further nodes are pushed away from the potential field (light yellow node in the Figure). The sampled set of poses will be denoted as $wp = \{wp^1,...,wp^m\}$ and is used to generate the problem description.   
   
%

Once a task action $a$ is selected for expansion from the current state $s_k$, the corresponding motion plan $q_{k_0},...,q_{k_n}$ is generated using a combination of planner discretization and a constant distance factor that will be detailed in Section~\ref{subsection:External}. At each $q_{k_j}$, the PDDL+ event \textit{belief\_update} triggers the semantic attachment call to the external library. The external library performs the belief updates (\ref{eq:predict})-(\ref{eq:update}) attaching to the event effects the updated belief estimate. The returned semantic attachment effects guides the staged RPG (SRPG) construction. Consequently, the belief estimate returned by the semantic attachments guide the SRPG in identifying the \textit{helpful actions}, besides providing an efficient heuristic evaluation. A weighted $\textrm{A}^\star$ forward state space search guides the state graph building phase, synthesizing a plan that minimizes the objective function (\ref{eq:cost}).  



\subsection{External Calls in PDDL+}
\label{subsection:External}

\begin{figure}[t!]
        \centering
        \includegraphics[width=1\textwidth]{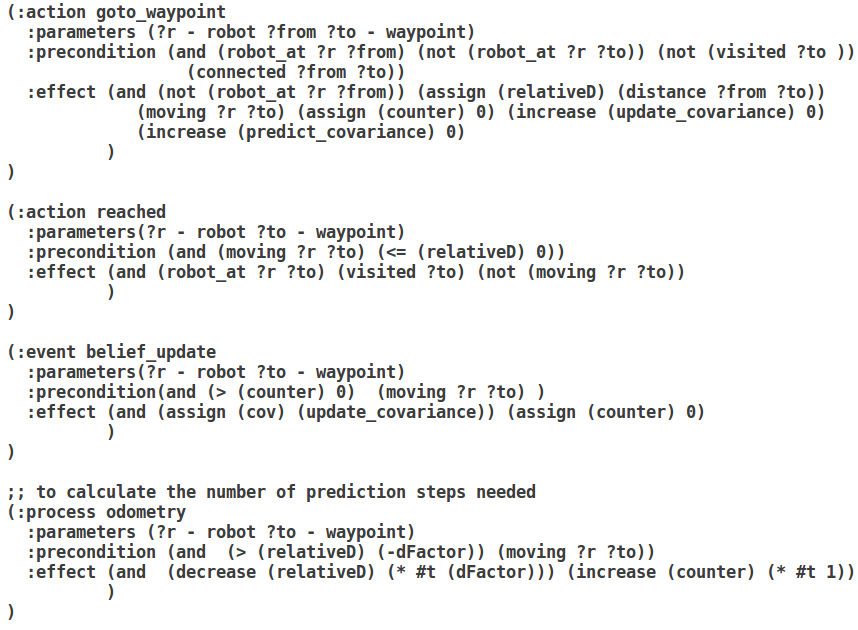}
        \caption{A fragment of the mobile robot domain with the precess and event. The \textit{process} odometry is used to simulate the robot translation and the \textit{event} belief\_update performs the belief propagation and posterior computation using semantic attachments.}
 \label{fig:domain}
\end{figure}

A snippet of the PDDL+ description for our mobile robot scenario is shown in Fig.~\ref{fig:domain}. From any state $s_k$, once the action \textit{goto\_waypoint} is triggered, the corresponding motion plan $q_{k_0},...,q_{k_n}$ is simulated by the process \textit{odometry} with the control $u_k$ divided into $n$ discrete translations of length $\delta_{trans_k} = \Delta \times dFactor$. DiNo is based on the planning-via-discretization approach and $\Delta$ is the user defined planner discretization, which together with the user specific motion discretization constant $dFactor$ determines the discretization $n$ for the control $u_k$. The event \textit{belief\_update} then updates the belief estimates for each of the $q_{k_j}$ configuration. In this way, we extend the DiNo planner to incorporate semantic attachments, computing the propagated belief $b[\bar{X_{k+1}}]$ upon executing a control $u_k$ at state $x_k$, as well as the posterior belief $b[X_{k+1}]$ upon obtaining a measurement. Since we are in the planning phase and yet to obtain observations, we simulate future observations $z_{k+1}$ given the propagated belief $b[\bar{X_{k+1}}]$, the set of landmarks $lm$ and the measurement model (\ref{eq:measurement_model}). Given a pose $x \in b[\bar{X_{k+1}}]$, the nominal observation $\hat{z} = h(x, lm_i)$ is corrupted with noise to obtain $z_{k+1}$.

The formal algorithm for performing the belief updates is summarized in Algorithm~\ref{algo:EKF_PDDL}. The focal elements in the planning domain are: the action \textit{goto\_waypoint}, the event \textit{belief\_update} that triggers the external call to evaluate to perform the belief updates and the process \textit{odometry} that simulates the robot motion between each planner discretization $\Delta$. Starting from a given pose (line~\ref{in:algo:action1}) the \textit{goto\_waypoint} action (line~\ref{in:algo:action2}) initiates the robot motion towards a connected pose. The immediate effect of this action is to initialize the distance between the two poses (line~\ref{in:algo:action3}) which starts the process \textit{odometry} as seen in line~\ref{in:algo:process}. The process effect simulates the translational motion at each $\Delta$, decreasing the distance between the poses by $\delta_{trans_k} = \Delta \times dFactor$ (line~\ref{in:algo:disctretize}). Event \textit{belief\_update} is immediately initiated (line~\ref{in:algo:predict}) which triggers the semantic attachment call to perform belief update, returning the propagated belief $b[\bar{X_{k+1}}]$ to the event effect (line~\ref{in:algo:predict}). If a landmark $lm_i \in lm$ is within the sensor range, the posterior belief $b[X_{k+1}]$ is evaluated returning its trace (line~\ref{in:algo:update}) to the event effect. To ensure the \textit{process-event-process} ordering we employ a variable $counter$ as shown in Fig.~\ref{fig:domain}.   

%

\begin{algorithm}[t!]
\label{alg:problem statement}
\caption{Belief update implementation in PDDL+}
\label{algo:EKF_PDDL}
\begin{algorithmic}[1]
\Require{Set of poses $wp$, set of landmarks $lm$, trace of initial pose covariance $trace(\Sigma_0)$, motion discretization $dFactor$, PDDL+ process discretization $\Delta$}
\While{ $\neg $(robot\_at $goal pose$)}
\State{(robot\_at $wp\_{from}$)}
\label{in:algo:action1}
\State{action $goto\_waypoint$}
\label{in:algo:action2}
\State{$d(from, to)$ = distance($wp\_{from}$, $wp\_{to}$ )}
\label{in:algo:action3}
\While{$d(from,to) > - \delta_{trans_k}$}
\State{process $odometry$}
\label{in:algo:process}
\State{$d(from,to) \leftarrow d(from,to)- $ $\delta_{trans\_k}$}
 \label{in:algo:disctretize}
\State{event $belief\_update$}
\State{$trace(\bar{\Sigma}_{k+1})$ $\leftarrow$ $trace(\Sigma_k$)}  \Comment{Eq.~\ref{eq:predict}}
\label{in:algo:predict}
\If{$\textrm{landmark within sensor range}$}
 \State{$trace(\Sigma_{k+1})$ $\leftarrow$ \texttt{$trace(\bar{\Sigma}_{k+1})$}} \Comment{Eq.~\ref{eq:update}}
\label{in:algo:update}
\EndIf
\EndWhile
\State{:action $reached$}
\State{(robot\_at $wp\_{from}$) $\leftarrow$ (robot\_at $wp\_{to}$)} 
\EndWhile
\State \Return{plan}
\end{algorithmic}
\end{algorithm}


%% file: results.tex
\label{section:results}
In this Section, we evaluate our approach in a simple yet realistic scenario in the Gazebo simulator. The performance are evaluated on an Intel{\small\textregistered} Core i7-6500U under Ubuntu 16.04 LTS.

\begin{figure}[]
	\centering
	
	\subfloat{\includegraphics[scale=0.15]{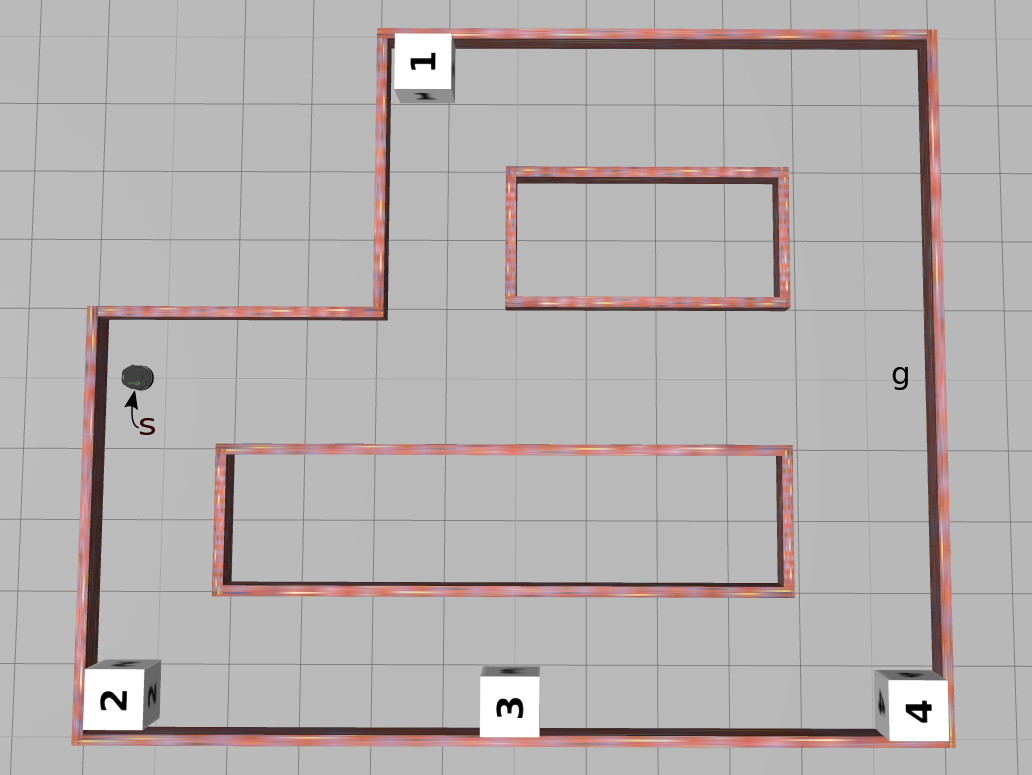}\label{fig:gazebo_new}}\\
 \vspace{-0.25cm}
\subfloat{\includegraphics[ scale=0.129]{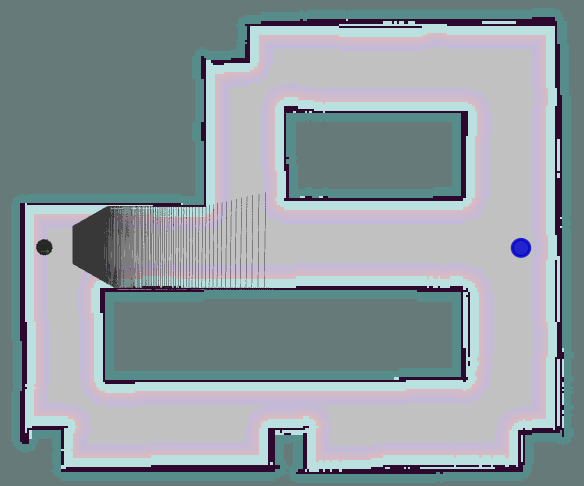}\label{fig:starting_config}}
	\hspace{0.03pt}
\subfloat{\includegraphics[ scale=0.127]{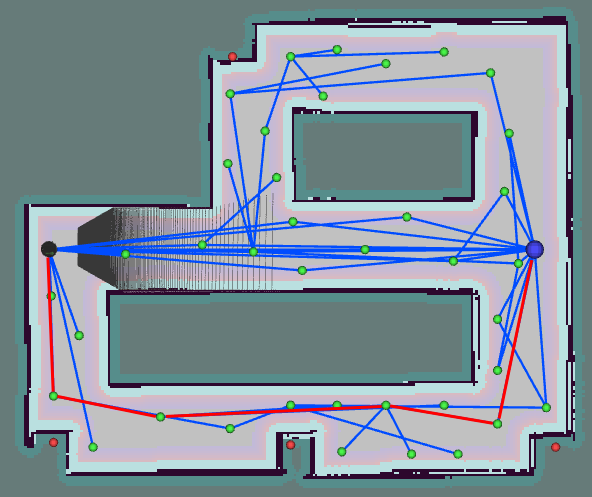}\label{fig:case4}}
		\caption{(\textit{top}) Considered scenario in gazebo. (\textit{bottom left}) Mapped environment. (\textit{bottom right}) Sampled poses and the planned trajectory (in red) for a particular scenario. \label{fig:environemnt}}
\end{figure}

Consider the corridor environment as seen in Fig.~\ref{fig:environemnt} (top), where the turtlebot robot starting from the initial pose ($s$ in Figure) needs to reach the goal pose ($g$ in Figure) to recharge its batteries. The turtlebot is initially oriented towards $g$. The cubes marked 1-4 are the landmarks in environment. The \textit{slam\_gmapping} ROS package is used to build the environment map. The resulting map of the environment is shown in Fig.~\ref{fig:environemnt} (bottom left). The turtlebot with its laser scanner can be seen facing the goal pose (blue in Figure). 

To begin with, poses are sampled according to the approach mentioned in Section~\ref{subsec:workflow}. 
While the planning time scales exponentially with the number of poses, the SRPG based search reduces this state space explosion significantly. Furthermore, due to our potential field based RRT sampling, we are able to prune unwanted state expansions by generating parsimoniously connected poses which are sufficient for synthesizing satisficing plans. We also discretize the distance between two connected poses using by a factor $\delta_{trans_k}$ and thereby the number of times the belief is updated; also regulating the planning time. 


\begin{figure}[t!]
	\centering
	
	\subfloat[$\Delta = 0.5$]{\includegraphics[scale=0.27]{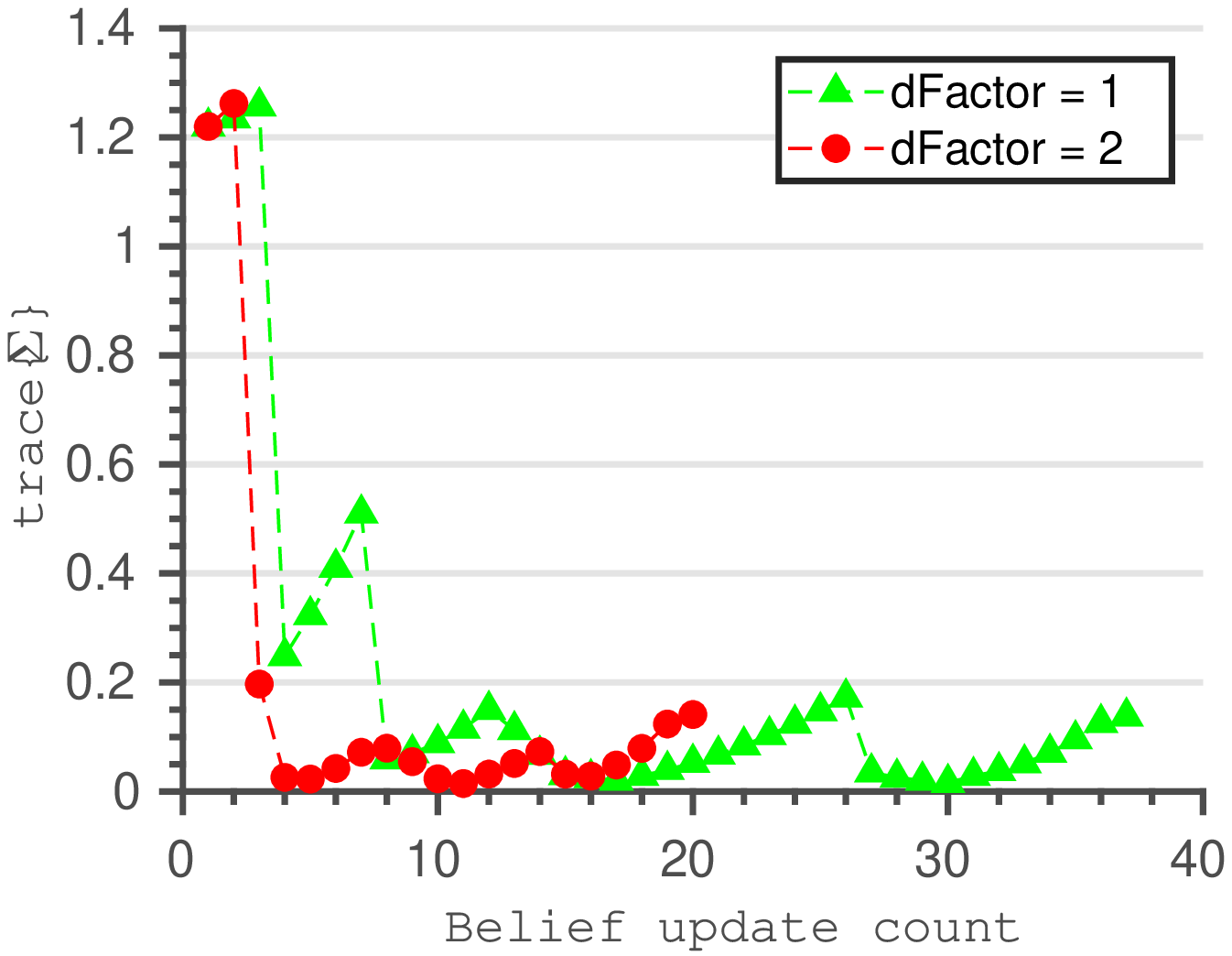}\label{fig:d=0,5}} 
 	\hspace{2pt}
	\subfloat[$\Delta=1$]{\includegraphics[scale=0.27]{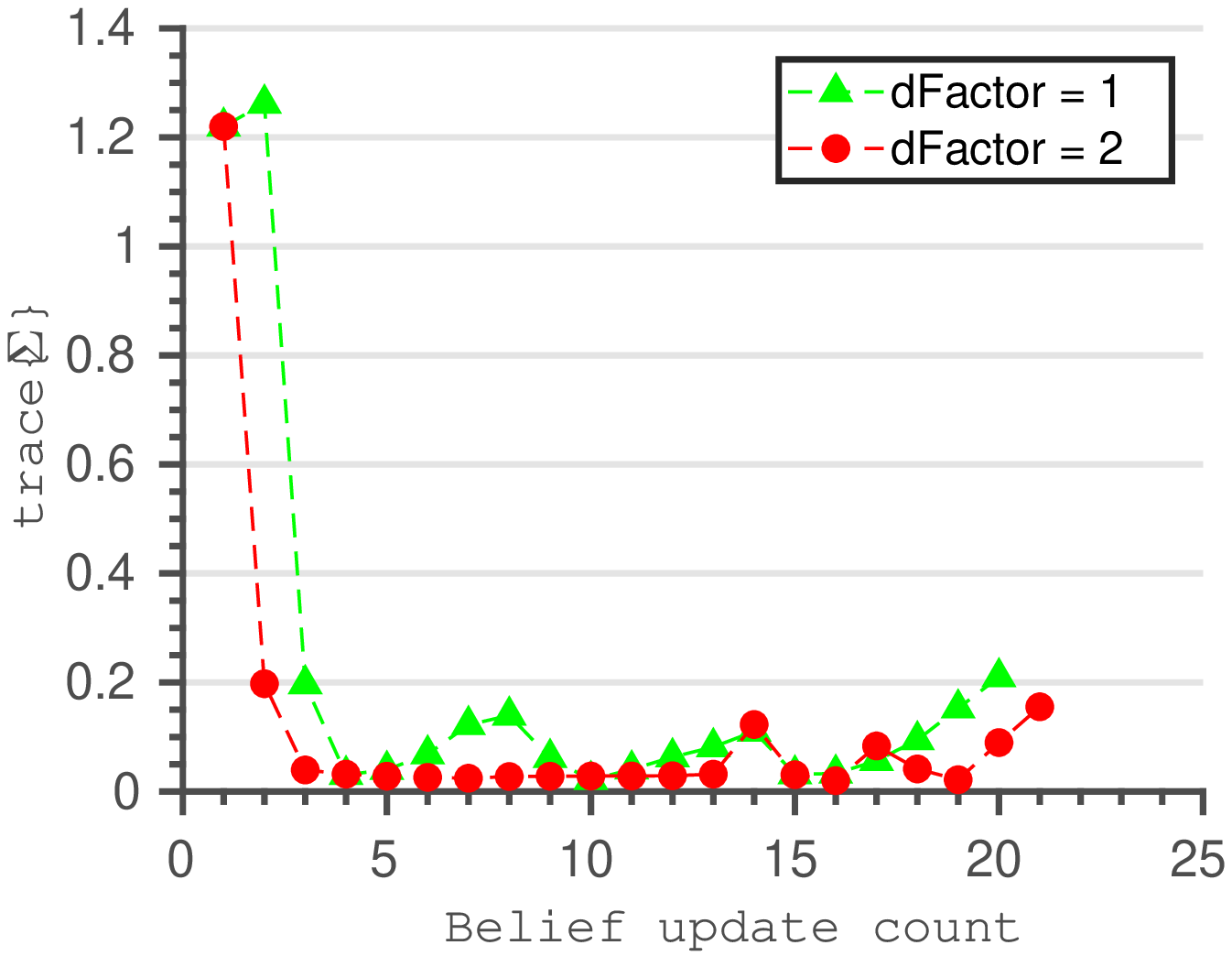}\label{fig:d=1}} 
	\hspace{2pt}
		\subfloat[$\Delta=2$]{\includegraphics[scale=0.27]{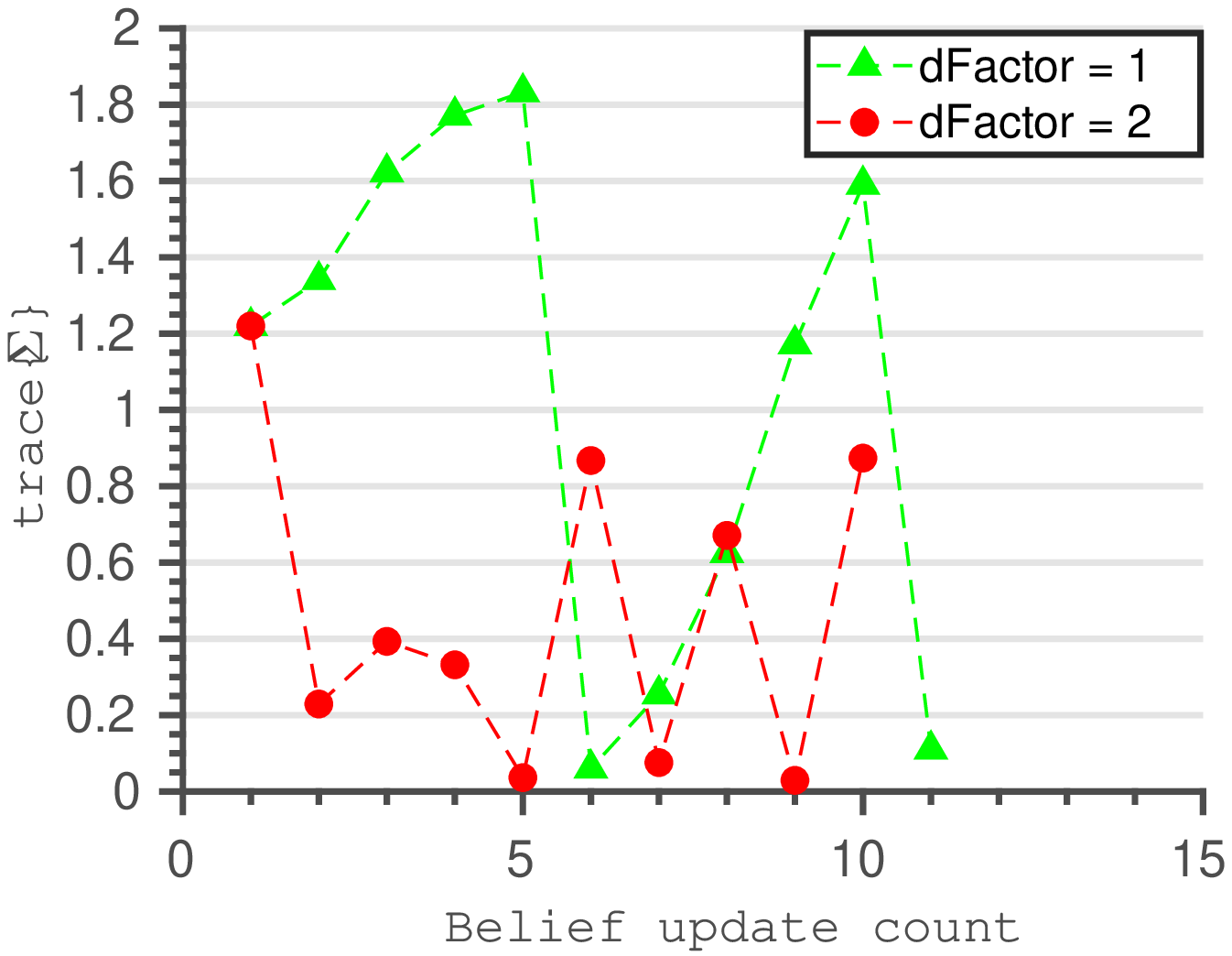}\label{fig:d=2}}
		\hspace{2pt}
		\subfloat[$\Delta=3$]{\includegraphics[scale=0.27]{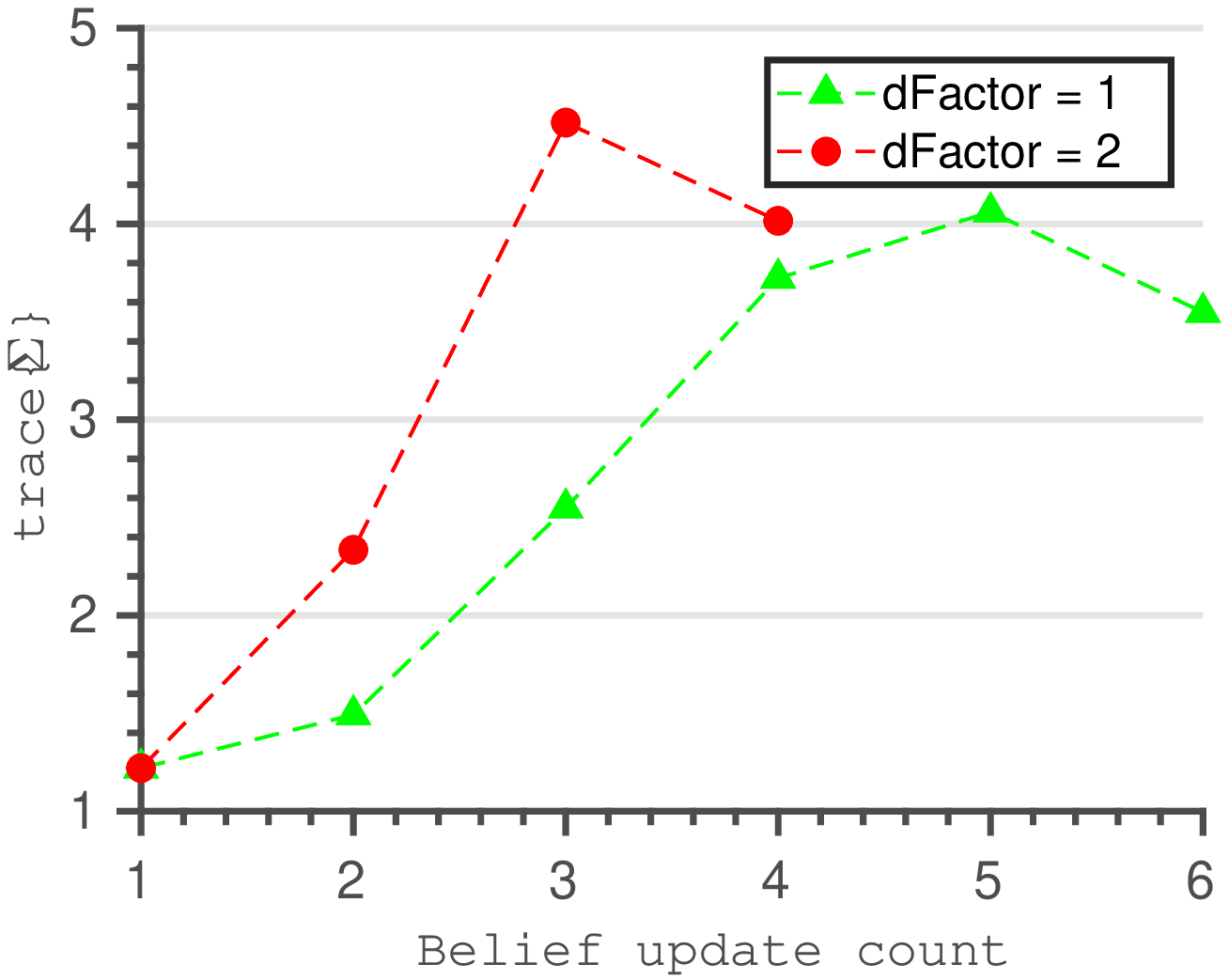}\label{fig:d=3}}
		\hspace{2pt}
		\caption{Covariance trace evolution for each \textit{belief\_update} event with different values of planner discretization.\label{fig:evolution}}
\end{figure}

The results for different discretization are summarized in Table~\ref{table:discretization}. As discussed before, DiNo approximates the continuous system dynamics using discretized uniform time steps $\Delta$. Hence, the plans synthesized are ratified against the continuous model using VAL~\cite{howey2004ICTAI}. A coarse $\Delta$ can lead to skipping certain decision points and thereby producing valid but flawed plans. This can be noticed from Fig.~\ref{fig:d=2} and~\ref{fig:d=3}. Particularly, in Fig.~\ref{fig:d=2} for $dFactor = 1$, $trace(\Sigma_g)$ is found to be the lowest. However, it is quite evident that the robot might collide with the walls, due to the erratic uncertainty evolution.

\begin{table}[ht] 
\centering
\scalebox{0.9}{
\begin{tabular}{|c| c c c c|}
\hline
dFactor & $\Delta$ & States explored & $t$ (s) &  $trace\{\Sigma_g\}$\\
\hline
\multirow{4}{*}{1} 
& 0.50 & 71751  & 3065.00 & 0.14\\
& 1.00 & 19407 & 549.96 & 0.21 \\
& 2.00 & 2486 & 46.48 & 0.11\\ 
& 3.00 & 218 & 3.74 & 3.55\\ \hline

\multirow{4}{*}{2} 
& 0.50 & 98659  & 4373.26 & 0.14\\
& 1.00 & 6675 & 144.38 & 0.16 \\
& 2.00 & 546 & 10.14 & 0.87 \\ 
& 3.00 & 162 & 2.90 & 4.02\\ \hline
\end{tabular}}
\caption{Analyzing TMP for different $dFactor$ and $\Delta$. $t$ denotes the total planning time. All results are for temporal horizon $T = 20$ and $m=40$ poses.} 
\label{table:discretization}
\end{table}

A finer discretization improves accuracy but at the expense of increased state space. This is clearly observable in Table~\ref{table:discretization}, from the number of states explored and the time taken for $\Delta = 0.5$. Fig.~\ref{fig:d=0,5} and~\ref{fig:d=1} show similar trend in the covariance evolution and hence considering the planning time, the optimal values for $\Delta$ and $dFactor$ are $1.00$ and $2$ respectively with about 145 seconds for planning. It is true that higher values for $dFactor$ would reduce the planning time. However, the selection of this value depends on the minimum distance between the poses.

\begin{figure}[h]
	\centering
	\caption{Snippet showing the modeling of battery charge drop rate.}
	\includegraphics[scale=0.32]{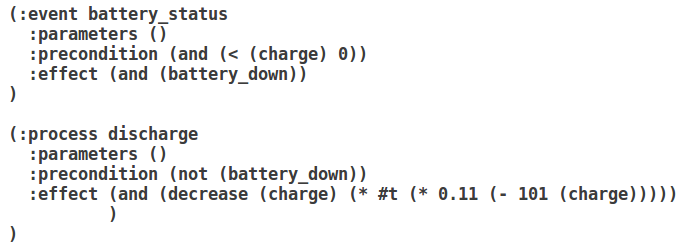}
	\label{fig:battery}
\end{figure}

To corroborate with the motivating example in Section~\ref{intro}, the domain in Fig.~\ref{fig:domain}, is extended to include a process \textit{discharge} to discharge the battery at a linear rate, and an event \textit{battery\_status} to check if the battery is down (see Fig.~\ref{fig:battery}). The kobuki platform works on 12 volts and 1.5 ampere, in direct connection with the laptop battery. Using a standard laptop battery of 43WHR lithium ion battery, the platform will function for 2.5 hours which approximates to a linear discharge rate of 0.011. To illustrate our approach, we assume a simplistic yet pragmatic rate of $d(\textrm{charge})/dt = -0.11(101 - \textrm{charge})$, where $charge$ is the percentage battery charge remaining.     

The goal condition is now modified to include the fact $\neg(battery\_down)$. In addition we also add an supplementary goal condition, trace$(\Sigma_g) < \eta$, where $\eta$ is a constant. Many practical applications require such bounds, for example, a charging cable of short length would mean that, there is a bound on the maximum pose uncertainty the robot can afford upon reaching $g$. Starting with a charge of 80$\%$ and $\eta = 0.20$, a plan is found in about 90 seconds, with 5143 state expansions, giving trace$(\Sigma_g) = 0.18$. However, starting with 40$\%$ charge and the same value of $\eta$, no valid plan exists and the planner ran out of memory upon expanding 140000 states. This manifests the fact that considering the motion plan alone can be catastrophic, as in the above scenario, the robot would have stopped in between had it executed the motion plan alone without logically reasoning regarding the battery constraint.


%% file: conclusion.tex
We have discussed an approach for integrated Task-Motion Planning for navigation, equipping a hybrid task planner with the capability of reasoning in the belief space of the robot. Collision-free configurations are sampled and the task planner synthesizes plans by directly planning over the sampled set. Expressive power of PDDL+ combined with heuristic based semantic attachments simulate the belief evolutions given an action sequence and the corresponding expected future observations. The underlying methodology of the hybrid planner has been discussed, validating the approach using a realistic synthetic scenario developed in Gazebo. Our sampling strategy combined with motion discretization help reduce the state space explosion while planning. To the best of our knowledge no other TMP approach utilizes the expressive power of PDDL+, which enables encompassing motion planning within the task domain. 

While the scalability to larger domains still remains a challenge, exploiting the planning-as-model-checking nature of the DiNo planner along with efficient caching of plans might help in tackling this issue to some extent. Effective sampling strategies can help in pruning the unwanted state expansion. The extant of such pruning needs to be studied in detail. Currently we perform off-line planning and an immediate future work include extending to plan-infer-plan paradigm. We also plan to validate our approach using the well-known benchmark problems~\cite{lagriffoul2018RAL}.

